\documentclass[runningheads]{llncs}
\usepackage{makeidx}
\usepackage{amsmath}
\usepackage{mathtools}
\usepackage[colorlinks]{hyperref}

\usepackage{floatrow}
\newfloatcommand{capbtabbox}{table}[][\FBwidth]
\usepackage{booktabs}
\usepackage{amssymb}

\title{The Minimum Cost Connected Subgraph Problem in Medical Image Analysis}
\titlerunning{The Minimum Cost Connected Subgraph Problem}

\author{
  Markus Rempfler\inst{1} \and
  Bjoern Andres\inst{2} \and
  Bjoern H.\ Menze\inst{1}}

\authorrunning{M.\,Rempfler et al.}

\institute{
  Institute for Advanced Study \& Department of Informatics, \\ Technical University of Munich, Germany \and
  Max Planck Institute for Informatics, Saarbr\"ucken, Germany}

\usepackage{bbold}      
\usepackage{mathtools}  

\DeclareMathOperator*{\argmax}{arg\,max}

\newcommand{\figref}[1]{Fig.~\ref{#1}}
\newcommand{\secref}[1]{Sec.~\ref{#1}}
\newcommand{\tableref}[1]{Table~\ref{#1}}

\newcommand{\ie}{i.e.\ }
\newcommand{\eg}{e.g.\ }

\renewcommand{\vec}[1]{\mathbf{#1}}

\newcommand{\neighbour}[2]{\delta^{#2}(#1)}

\newcommand{\separatorset}{\mathcal{S}}

\newcommand{\tree}[1]{T\left( #1 \right)}
\newcommand{\nodes}{V}
\newcommand{\vertices}{\nodes}
\newcommand{\edges}{E}
\newcommand{\x}{\vec{x}}
\newcommand{\X}{\vec{X}}
\newcommand{\Xx}{\X=\x}

\newcommand{\sol}{\x^*}

\newcommand{\feas}{\Omega}

\newcommand{\cp}[2]{P\left( #1\vert #2\right)}

\usepackage[acronym,shortcuts,nonumberlist]{glossaries}
\glsdisablehyper

\newacronym{mip}{MIP}{mixed integer programming}
\newacronym{map}{MAP}{maximum a posteriori}
\newacronym{ilp}{ILP}{integer linear program}
\newacronym{ip}{IP}{integer programming}
\newacronym{lp}{LP}{linear program}
\newacronym{mrf}{MRF}{Markov random field}
\newacronym{crf}{CRF}{conditional random field}
\newacronym{dsa}{DSA}{digital subtraction angiography}

\newacronym{bfs}{BFS}{breadth-first search}
\newacronym{dfs}{DFS}{depth-first search}

\newacronym{mccs}{MCCS}{minimum cost connected subgraph}

\newcommand{\ilp}{\acs{ilp}}
\newcommand{\map}{\ac{map}}

\begin{document}

\maketitle

\begin{abstract}
  Several important tasks in medical image analysis can be stated
  in the form of an optimization problem whose feasible solutions are
  connected subgraphs. Examples include the reconstruction of neural
  or vascular structures under connectedness constraints.

  We discuss the minimum cost connected subgraph (MCCS)
  problem and its approximations from the perspective of medical
  applications. We propose a)~objective-dependent constraints and
  b)~novel constraint generation schemes to solve this optimization
  problem exactly by means of a branch-and-cut algorithm. These are
  shown to improve scalability and allow us to solve instances of two
  medical benchmark datasets to optimality for the first time. This
  enables us to perform a quantitative comparison between exact and
  approximative algorithms, where we identify the geodesic tree
  algorithm as an excellent alternative to exact inference on the
  examined datasets.
\end{abstract}

\section{Introduction}

\begin{figure}[t]
  \centering
  \includegraphics[width=0.325\textwidth, trim=0.3cm 0 0.6cm 0,
  clip]{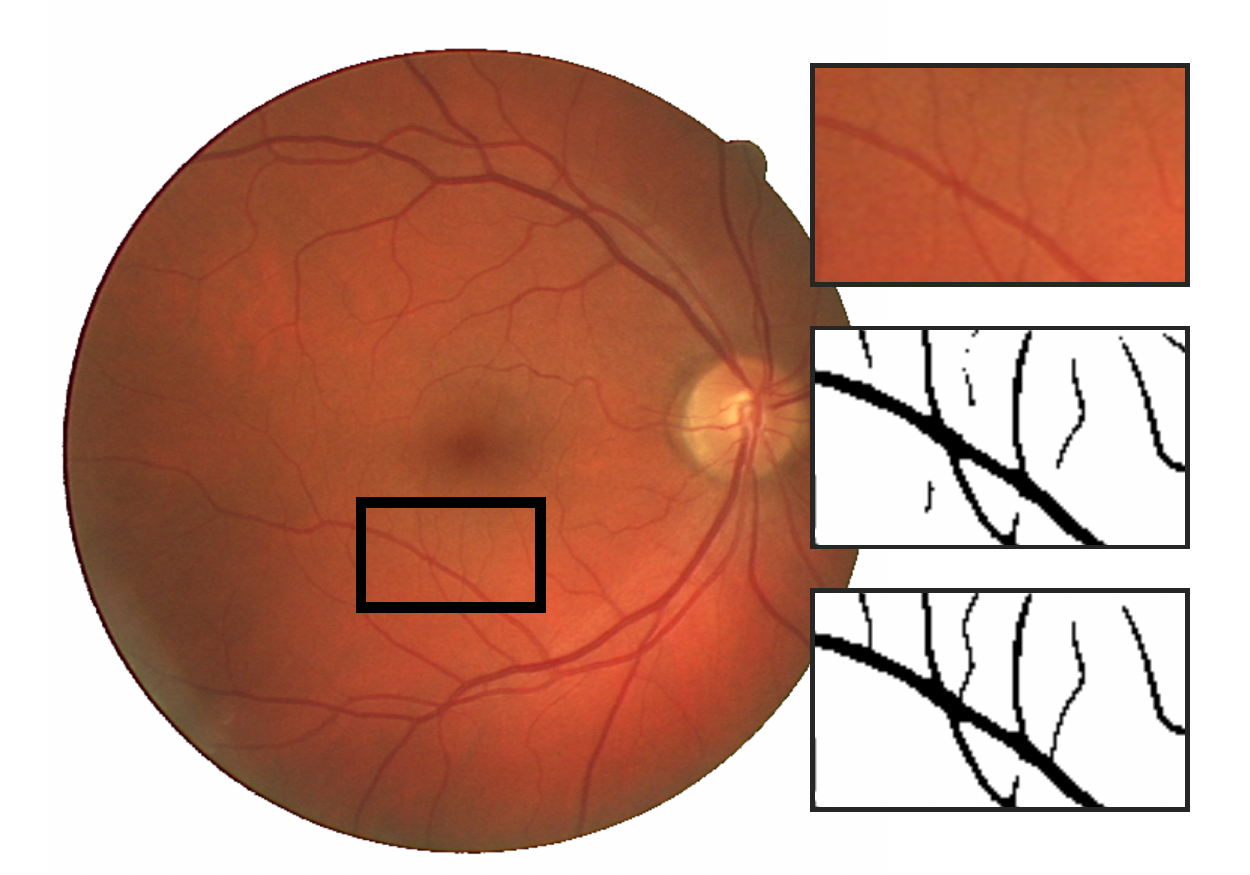}
  \includegraphics[width=0.325\textwidth, trim=0.3cm 0 0.6cm 0,
  clip]{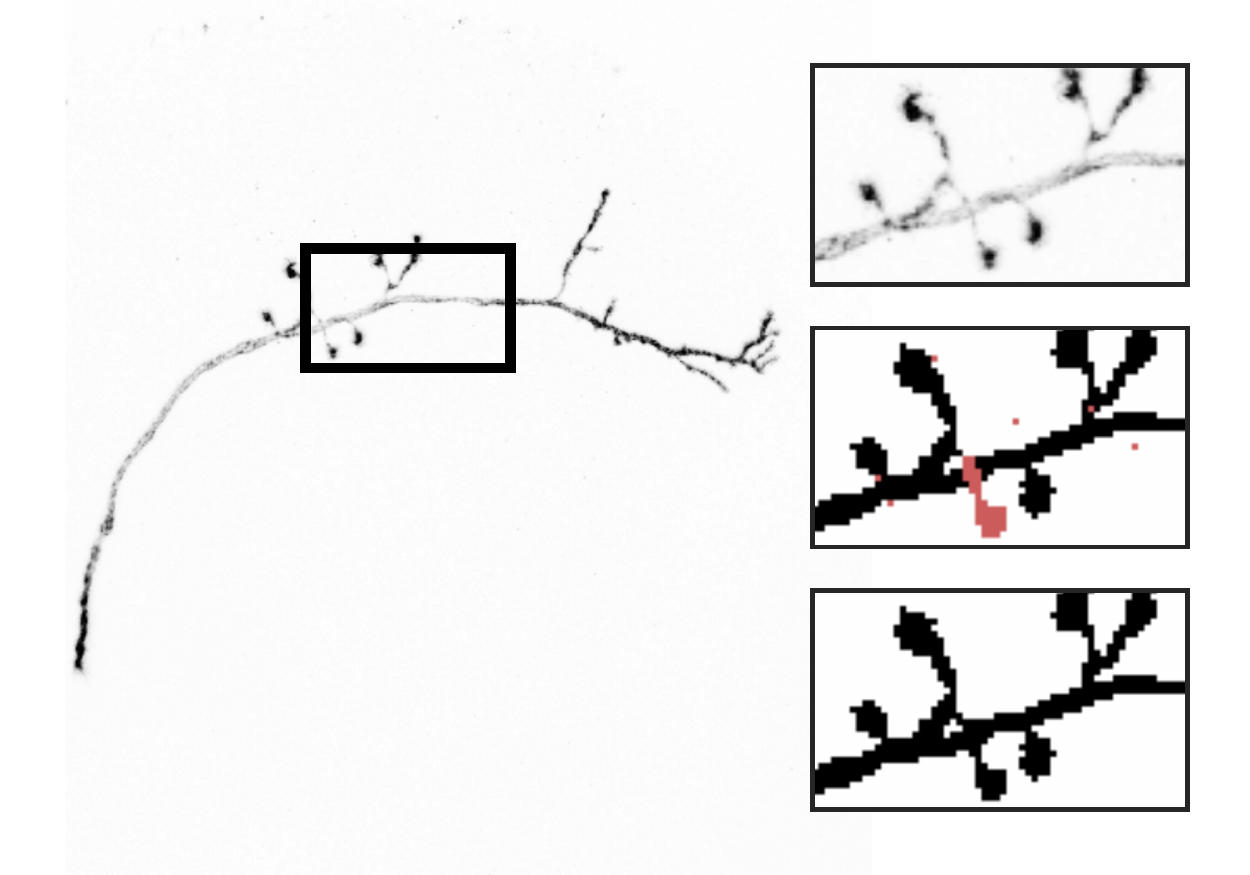}
  \includegraphics[width=0.325\textwidth, trim=0.3cm 0 0.6cm 0,
  clip]{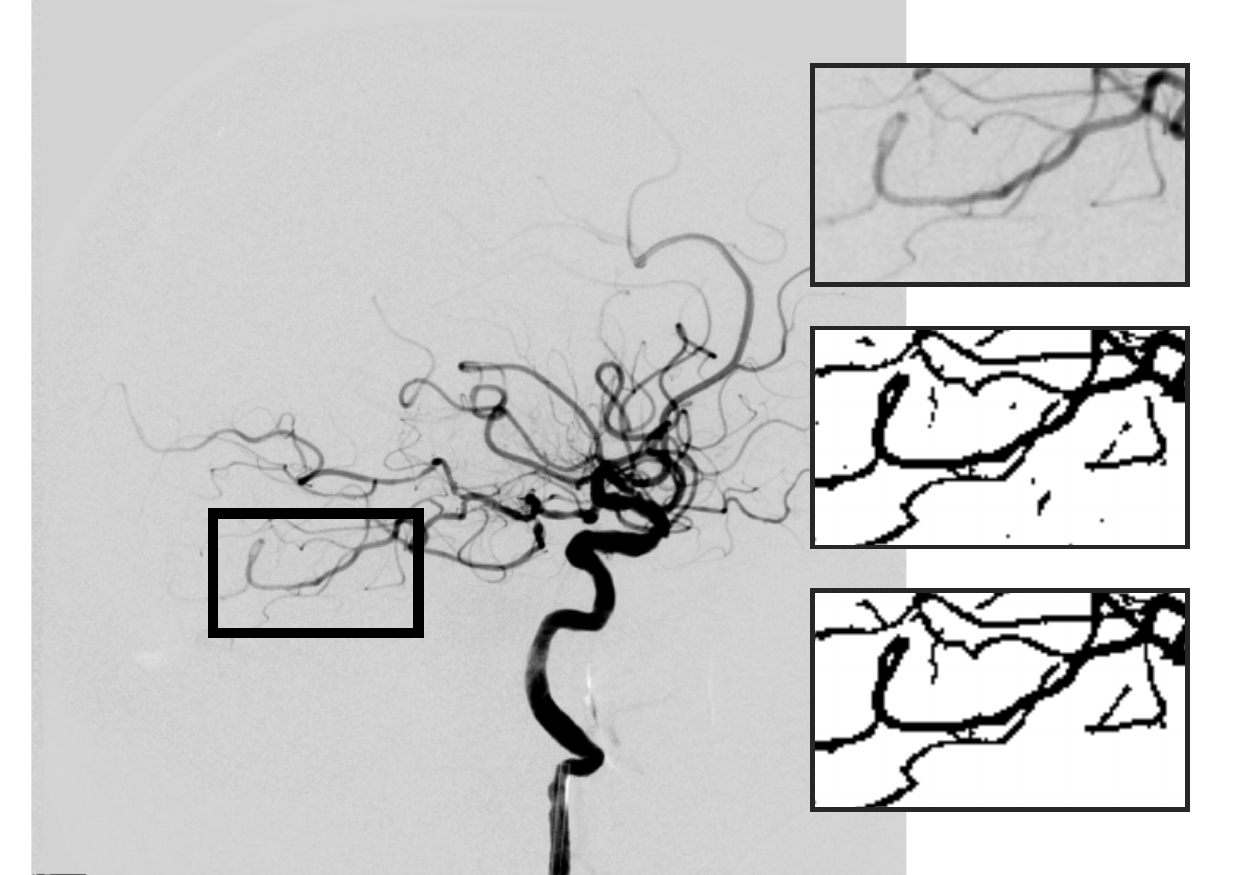}
  \caption{\textit{Examples for the \acs{mccs} on grid graphs}. {\bf
      Left}: Segmentation of vasculature in retinal images. {\bf
      Middle}: Reconstruction of a neuron from a 3D stack. Excessive
    disconnected components are shown in red for better
    visibility. {\bf Right}: Delineation of vessels in a digital
    subtraction angiography (DSA) time series. The detail views show:
    raw image (\textbf{top}), without connectedness (\textbf{middle})
    and with connectedness (\textbf{bottom}). Imposing connectedness
    constraints, \ie requiring an \ac{mccs}, helps to reconnect
    disconnected terminals and remove spurious detections without
    penalizing thin tubular structures.}
  \label{fig:teaser}
\end{figure}

The \emph{minimum cost connected subgraph} (MCCS) optimization problem
arises in several medical image analysis tasks, most prominently for
segmenting neural structures~\cite{Turetken2016} or reconstructing
vascular networks~\cite{Rempfler2015media}, where the \map{} subgraph
under connectedness constraints is inferred. Variations of this
optimization problem have been proposed for anatomical labelling of
vasculature~\cite{Robben2016} or artery-vein
separation~\cite{Payer2016}. Imposing connectedness serves as
regularizer, suppressing spurious detections and complementing
incomplete observations, and it is often a requirement for further
processing steps, \eg if the reconstructed vasculature shall be used
for biophysical simulations.

While~\cite{Turetken2016,Rempfler2015media,Robben2016,Payer2016}
successfully solve an \acs{mccs} problem on heavily preprocessed,
application-specific, sparse graphs, it would also be interesting to
enforce connectedness on both very dense or large grid-graphs, for
example in low-level segmentation tasks (\figref{fig:teaser}, left),
for 3D/4D reconstruction problems (\figref{fig:teaser}, middle and
right) or when it is not possible to reliably reduce the candidate
graphs size. In these cases, however, the computational complexity
becomes challenging. In fact, it was shown to be NP-hard
in~\cite{Vicente2008}. Nowozin \& Lampert~\cite{Nowozin2009} propose
an exact algorithm that tightens an outer polyhedral relaxation of the
connected subgraph polytope by cutting planes. However, without
guarantee to terminate in polynomial time, it was found to be too slow
to solve typical instances of medical benchmark datasets to
optimality. To this end, two heuristical algorithms were proposed by
Chen et al.~\cite{Chen2011} and St\"uhmer et
al.~\cite{Stuhmer2013}. They either use an approximative formulation
of the connected subgraph polytope by means of a precomputed geodesic
shortest path tree~\cite{Stuhmer2013} or iteratively solve a surrogate
problem that is based on altered weights of the original
problem~\cite{Chen2011}. Both approaches are fast enough for medical
applications and were reported to yield qualitatively promising
results. A quantitative comparison, however, has been prevented by the
prohibitively expensive computation of exact solutions to the
\acs{mccs} problem.

In this paper, we revisit the \ac{mccs} in an integer linear
programming (ILP) framework for \map{} estimation under connectedness
constraints. First, we contribute to the exact optimization by
proposing a) \emph{objective-dependent constraints} that reduce the
size of the polytope and hence, reduce the number of potential
solutions to explore, and b) \emph{constraint generation strategies}
beyond the standard \emph{nearest} and \emph{minimal} separator
strategy, which we show to have a strong impact on the runtime of the
\ilp{}. Both propositions together enable us to compute the \ac{mccs}
on several instances of two medical benchmark datasets -- addressing
vessel segmentation and neural fiber reconstruction -- to
optimality. Our second contribution is a first quantitative comparison
of the exact algorithm and the two heuristics in terms of runtime,
objective function and semantic error metrics.

\section{Background}

We are interested in the most likely binary labeling $\x \in \{0,
1\}^{\vert \vertices \vert}$ of the nodes $\vertices$ in the graph
$G=(\vertices, \edges)$. A node $i$ is active if $x_i=1$. By imposing
connectedness constraints, \ie $\x \in \feas$, the \map{} estimate
becomes a \acs{mccs} problem:
\begin{equation}
  \label{eq:problem}
  \sol = \argmax_{\mathclap{\x \in \{0,1\}^{\vert \nodes \vert}}} \cp{\Xx}{I, \feas}
  = \argmax_{\x \in \feas} \cp{\Xx}{I} \enspace ,
\end{equation}
where $I$ is the image evidence and $\feas$ denotes the set of $\x$
that are connected subgraphs of $G$. In this section, we discuss two
formulations of $\feas$, the exact formulation that
follows~\cite{Nowozin2009} and the geodesic tree formulation
of~\cite{Stuhmer2013}.

\subsection{Exact Connectedness}
\label{sec:exactcon}

Following~\cite{Nowozin2009}, we can describe $\feas$ with the
following set of linear inequality constraints
\begin{equation}
  \label{eq:subgraphpoly}
  \forall i,j \in \vertices, (i,j) \notin \edges : \forall \separatorset \in S(i,j) \quad x_i + x_j - 1 \leq \sum_{k \in \separatorset} x_k \enspace ,
\end{equation}
where $\separatorset$ is a set of vertices that separate $i$ and $j$,
while $S(i,j)$ is the collection of all vertex separator sets for $i$
and $j$. In other words, if two nodes $i$ and $j$ are active, then
they are not allowed to be separable by any set of inactive
nodes. Thus, a path of active nodes has to exist. In practice, this
set of constraints is too large to be generated in advance. However,
given a labelling $\x$ we can identify at least a subset of the
violated connectedness constraints in polynomial time, add them to the
\ilp{} and search for a new feasible solution. This approach is known
as \emph{lazy constraint generation}. In \secref{sec:strats}, we
detail on identifying and adding these constraints.
\\[2ex] \textbf{Rooted case.}
In many medical segmentation problems,
it is reasonable to assume that a root node can be identified
aforehand with an application-specific detector, manually or by a
heuristic, such as picking the strongest node in the largest
component. If a known root $r$ exists, it suffices to check
connectedness to the root node instead of all pairs of active
nodes. The constraints in \eqref{eq:subgraphpoly} then become
\begin{equation}
  \label{eq:rootedsubgraph}
  \forall i \in \vertices \setminus \{r\}, (r,i) \notin E : \forall \separatorset \in S(i,r) \quad x_i \leq \sum_{k \in \separatorset} x_k \enspace .
\end{equation}

\subsection{Geodesic Tree Connectedness}
\label{sec:geodesictreecon}

Alternative to the exact description of all connected subgraphs that
we discussed in the previous section, we can formulate a connectedness
prior as in~\cite{Stuhmer2013} on a \emph{geodesic shortest path tree}
$\tree{G}=(V, A \subseteq \edges)$ rooted in $r$. Here, $\tree{G}$ is
precomputed based on the unary potentials, \ie with edge weights
defined as $f(i,j) = \frac{1}{2} \left( \max(w_i, 0) + \max(w_j, 0)
\right)$. The set of feasible solutions is then given by the
inequalities:
\begin{equation}
  \label{eq:geodesicconnectedness}
  \forall i \in V \setminus \{ r \}, (p, i) \in \tree{G} \quad x_i \leq x_{p} \enspace ,
\end{equation}
where $p$ is the parent of $i$ in the geodesic tree $\tree{G}$. With
this set of constraints, a node $i$ can only be active if his parent
$p$ in the geodesic tree is also active, thus connecting all active
nodes to the root $r$ along the branches of $\tree{G}$. Advantages of
this approach are that only $\vert \vertices \vert - 1$ constraints are
necessary to describe the set of feasible solutions and that the
relaxation is tight. On the other hand, the inequalities of
\eqref{eq:geodesicconnectedness} describe a strict subset of
\eqref{eq:rootedsubgraph}, unless $\tree{G} = G$. Hence it might
discard an optimal solution that is feasible in
\eqref{eq:rootedsubgraph}.

\section{Methods}

Given the probabilistic model $\cp{\Xx}{I}$ of \eqref{eq:problem} is a
random field over $G=(V,E)$, we can write its \map{} estimator $\sol =
\argmax_{\x \in \{ 0, 1\}} \cp{\Xx}{I, \feas}$ as an \ilp{}. We will
assume for the remaining part that $\cp{\Xx}{I}=\prod_{i \in
  \vertices} \cp{x_i}{I}$, leading to the \ilp{}:
\begin{eqnarray}
  \label{eq:generalilp}
  \mathrm{minimize} \quad &&\sum_{i \in \vertices} w_i x_i \enspace , \\
  \label{eq:conconstr}
  \mathrm{s.t.} \quad &&\x \in \feas \enspace , \\
  \label{eq:integconstr}
  &&\x \in \{ 0, 1\}^{\vert \vertices \vert} \enspace ,
\end{eqnarray}
where \eqref{eq:conconstr} are the connectedness constraints, \ie
either \eqref{eq:rootedsubgraph} or \eqref{eq:geodesicconnectedness},
\eqref{eq:integconstr} enforces integrality, and $w_i$ are the weights
that can be derived as $w_i = -\log
\frac{\cp{x_i=1}{I}}{1-\cp{x_i=1}{I}}$. Higher order terms of the
random field can be incorporated by introducing auxiliary binary
variables and according constraints as done
in~\cite{Rempfler2015media}. Note, however, that \cite{Chen2011}
reported problem instances with weak or no pairwise potentials -- as we are
addressing them here -- to be amongst the most difficult.

\subsection{Objective-dependent Constraints}
\label{sec:objdepconst}

Given the problem with unary terms, we observe that, for any connected
component $\mathcal{U} \subset \vertices$ composed of
\emph{unfavourable} nodes only, \ie $\forall i \in \mathcal{U}, w_i >
0$, it can only be active in the optimal solution if there are at
least two active nodes in its neighbourhood:
\begin{equation}
  \label{eq:nonleafgeneral}
  \forall i \in \mathcal{U} \quad 2 x_i \leq \sum_{\mathclap{j \in \cup_{k \in \mathcal{U}} \neighbour{k}{}\setminus \mathcal{U}}} x_j \enspace ,
\end{equation}
where $\neighbour{k}{}$ is the set of neighouring nodes to $k$.  In
other words, unfavourable nodes can not form a leaf in the optimal
solution (otherwise, removing the unfavourable nodes would give us a
better solution without loosing connectedness). In the special case of
$\vert \mathcal{U} \vert = 1$, we can add the constraint from the
beginning. This removes feasible solutions from $\feas$ that are a
priori known to be suboptimal, hence reducing the search space in the
optimization and making it unnecessary to add a large set of separator
inequalities.
\\[2ex]
\textbf{Higher-order weights.} Even though we only define
\eqref{eq:nonleafgeneral} for unary weights, it is possible to adapt
the constraint to higher-order models by changing the condition to
$w_i + \min_{j \in \neighbour{i}{}} w_{ij} > 0$, provided the pairwise
weights $w_{ij}$ are only introduced for neighbouring nodes $i,j$ such
that $(i,j) \in \edges$.

\subsection{Constraint Generation Strategies}
\label{sec:strats}

\begin{figure}[t]
  \centering
  \includegraphics[width=0.275\linewidth]{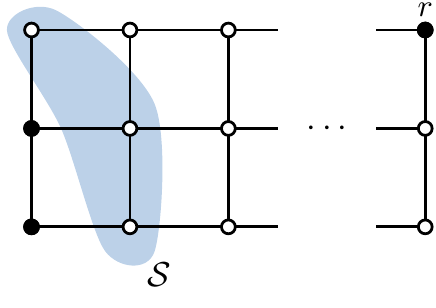}
  \hfill
  \includegraphics[width=0.275\linewidth]{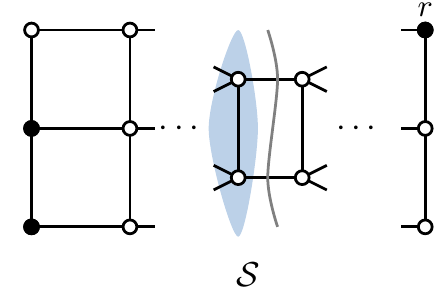}
  \hfill
  \includegraphics[width=0.275\linewidth]{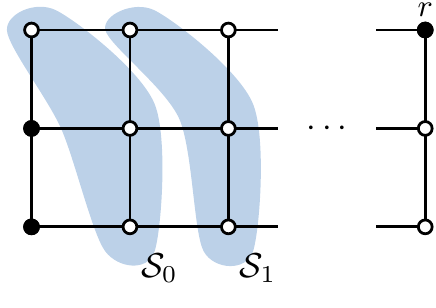}
  \caption{\emph{Constraint generation strategies}. Illustration of
    the nearest separator (left), minimal separator (middle) and
    $k$-nearest (right) strategies. Active nodes are shown in black,
    inactive nodes are white and the identified separator sets
    $\separatorset$ are marked in blue. $\separatorset$ is
    subsequently used to generate the corresponding constraint in
    \eqref{eq:subgraphpoly} or \eqref{eq:rootedsubgraph}.}
  \label{fig:strategies}
\end{figure}

The extensive number of inequalities needed for
\eqref{eq:rootedsubgraph} makes it necessary to identify violated
constraints during the optimization and add them to the problem. We
note that it suffices to treat individual connected components as one
entitity, since establishing a connection automatically connects all
pairs of nodes between them. Identifying violated constraints boils
down to finding a vertex separator set $\separatorset$ between two
disconnected, active components in the current solution. The
constraints corresponding to $\separatorset$ are then generated
according to \eqref{eq:subgraphpoly} or \eqref{eq:rootedsubgraph} for
all nodes in the given connected component.

At the heart of this technique is the observation that only a subset
of inequalities is active at the optimum of a given problem
instance. However, depending on the choice of the inequalities that we
add in each step, we may explore (and therefore construct) different
parts of the polytope $\feas$, most likely requiring a different
number of iterations.

In the following, we first review the two standard strategies, namely
the \emph{nearest} and \emph{minimal} separator, and then propose
several novel, alternative strategies.
\\[2ex]
\textbf{Nearest separator.} In this standard approach, the
vertex separator set in the immediate neighbourhood of the active
component is picked for generating the new constraint. This strategy
has been used, for example, in~\cite{Rempfler2015media}. It is
motivated by its simplicity and the fact that it often coincides with
the minimal separator strategy for small components.
\\[2ex]\textbf{Minimal separator.} A minimal (in terms of $\vert
\separatorset \vert$) separator set is obtained by solving a max-flow
problem between any two disjoint active components at hand and
selecting the smaller vertex set on either side of the resulting
min-cut. For the max-flow, we set the flow capacity $c$ in edge
$(i,j)$ as $c(i,j) = \max(1-x_i, 1-x_j)$. The strategy was applied
in~\cite{Nowozin2009}.
\\[2ex]\textbf{Equidistant separator.} Alternatively, we can identify
the separator set $\separatorset$ that is equidistant to the current
active component and all other components by running a \ac{bfs} from
either side. Similar to the max-flow of the minimal separator, the
distance measure is only accounting for non-active nodes. This
strategy originates in the observation that the weakest evidence
between two components is often found half-way into the connecting
path.
\\[2ex]\textbf{$k$-Nearest and $k$-Interleave.} We run a \ac{bfs} from
the active component $C$ and collect the $k$ (disjoint) separator sets
$\{ \separatorset_n \}_{n=0}^{k-1}$ composed of all nodes with
identical distance. The search terminates if $k$ equals the number of
nodes in $C$ or if another active node is reached. For the
$k$-interleave, only separators with even distance are chosen. The
intuition behind these strategies is that a wider range of neighbours
(and their neighbours) has to be considered for the next solution.

\section{Experiments \& Results}

\subsubsection{Datasets \& Preprocessing.}

We conduct experiments on two medical datasets: First, on the DRIVE
database of retinal images~\cite{Staal2004}, each being $565 \times
584\,\mathrm{px}$. We use the probability estimates $\cp{x_i=1}{I}$
for a pixel $i$ being vasculature from the recent state-of-the-art
approach of~\cite{Ganin2014} for our unaries. Second, we run
experiments on the olfactory projection fibers (OPF)
dataset~\cite{Brown2011}, composed of $8$ 3D confocal microscopy image
stacks. We use the stacks prepared in~\cite{Turetken2016}, where we
estimate $\cp{x_i=1}{I}$ of voxel $i$ being part of the fiber by a
logistic regression on the image intensities. We segment the nerve
fiber under the requirement of connectedness on the 3D grid graph of
$256\times256\times n$ nodes with $n \in \{ 30, \ldots, 51\}$
depending on the case. The probability $\cp{x_i=1}{I}$ of voxel $i$
being part of the fiber is estimated by a logistic regression on the
image intensities. Both datasets are illustrated in
\figref{fig:teaser}.

\subsubsection{Optimization.}
We solve the \ilp{} \eqref{eq:generalilp} by the branch-and-cut
algorithm of the solver Gurobi~\cite{Gurobi2015} with a default
relative gap of $10^{-4}$. Objective-dependent constraints for single
nodes (\secref{sec:objdepconst}) are added from the beginning. For the
exact connectedness (\secref{sec:exactcon}), the strategies described
in Sect.~\ref{sec:strats} are implemented as a callback: Whenever the
solver arrives at an integral solution $\x'$, violated constraints are
identified and added to the model. If no such violation is found, \ie
$\x'$ is already connected, then it is accepted as new current
solution $\sol$. For the geodesic tree connectedness
(\secref{sec:geodesictreecon}), all constraints are added at once. In
order to arrive at a fair comparison, we define the root node for both
approaches.

\subsubsection{Experiment: Objective-dependent constraints.}
\begin{figure}[t]
  \floatbox[{\capbeside\thisfloatsetup{capbesideposition={right,top},
capbesidewidth=sidefil}}]{figure}[\FBwidth]
{\caption{Runtime with and without the proposed objective-dependent
    constraints on $64 \times 64$ instances. Mean values are depicted
    by $\blacklozenge$, whiskers span $[\min,\max]$ values. Unsolved
    instances are excluded for readability. We find that all
    strategies benefit from the additional constraints. Additional
    \textit{per-instance} information can be found in the supplement.
    }
  \label{fig:objdep}}
{
 \includegraphics[width=0.36\textwidth]{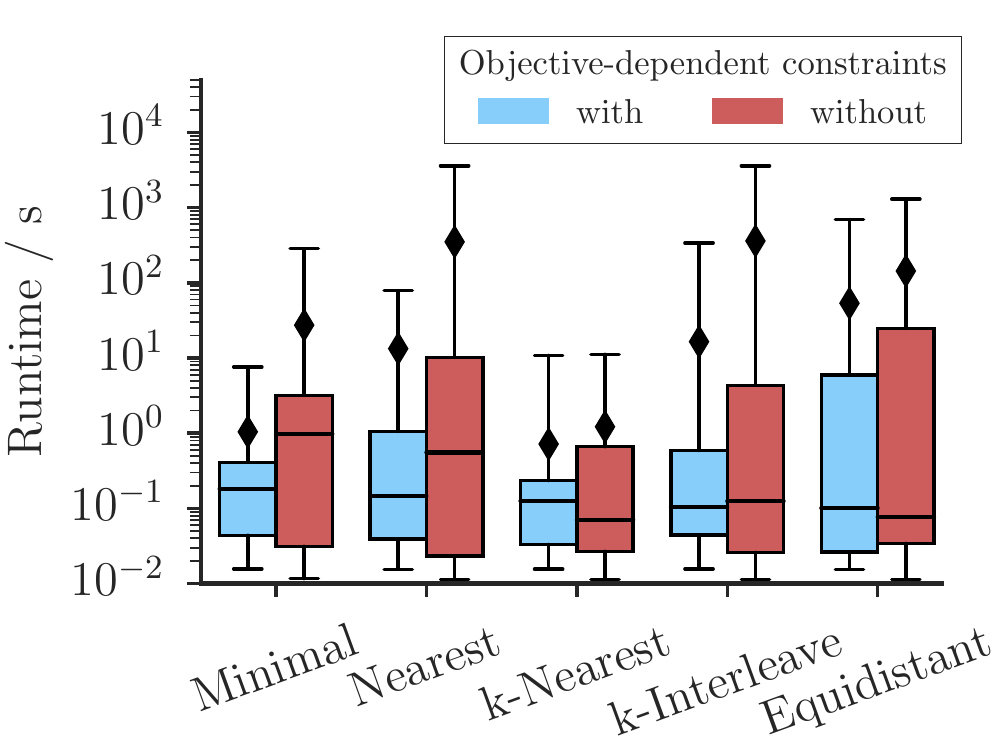}
}
\end{figure}

To examine the impact of the objective-dependent constraints, we
subsample 25 subimages of $64 \times 64\, \mathrm{px}$ from the DRIVE
instances and run the \ilp{} once with and once without the additional
first order constrains of \eqref{eq:nonleafgeneral}. As shown in
\figref{fig:objdep}, we find that all strategies benefit from the
additional constraints.

\subsubsection{Experiment: Comparing exact and approximative
  algorithms.}

\begin{figure}[t]
  \begin{floatrow}
    \centering \ffigbox[0.52\textwidth][]{
      \includegraphics[width=0.25\textwidth]{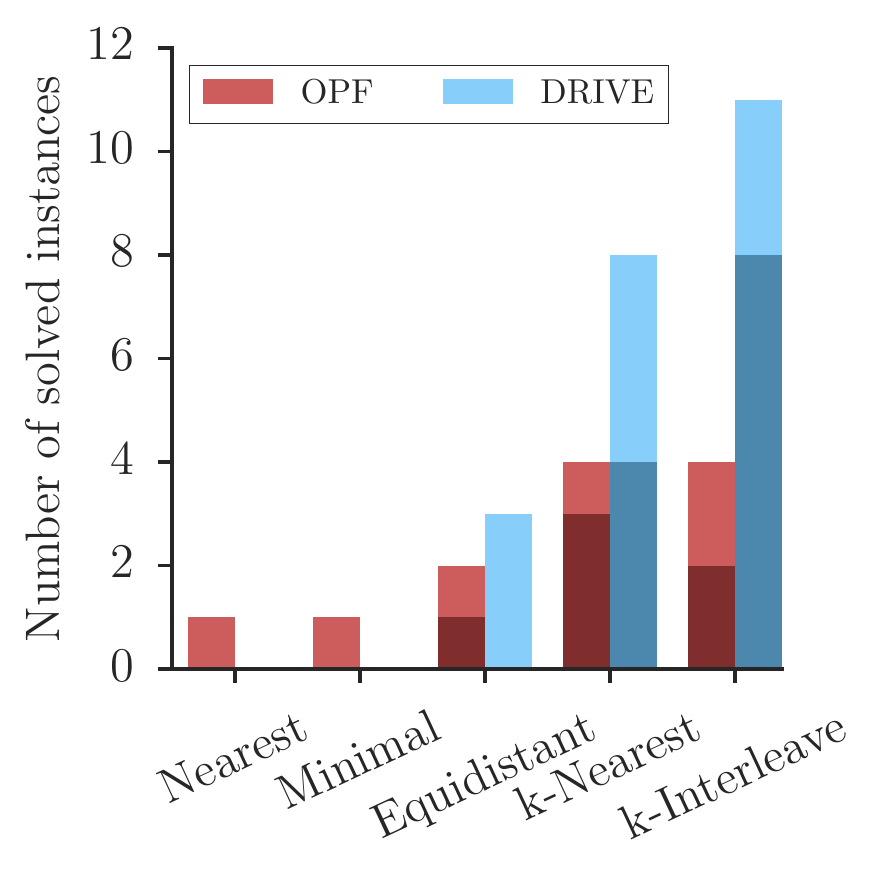}
      \hspace{-0.2cm}
      \includegraphics[width=0.25\textwidth]{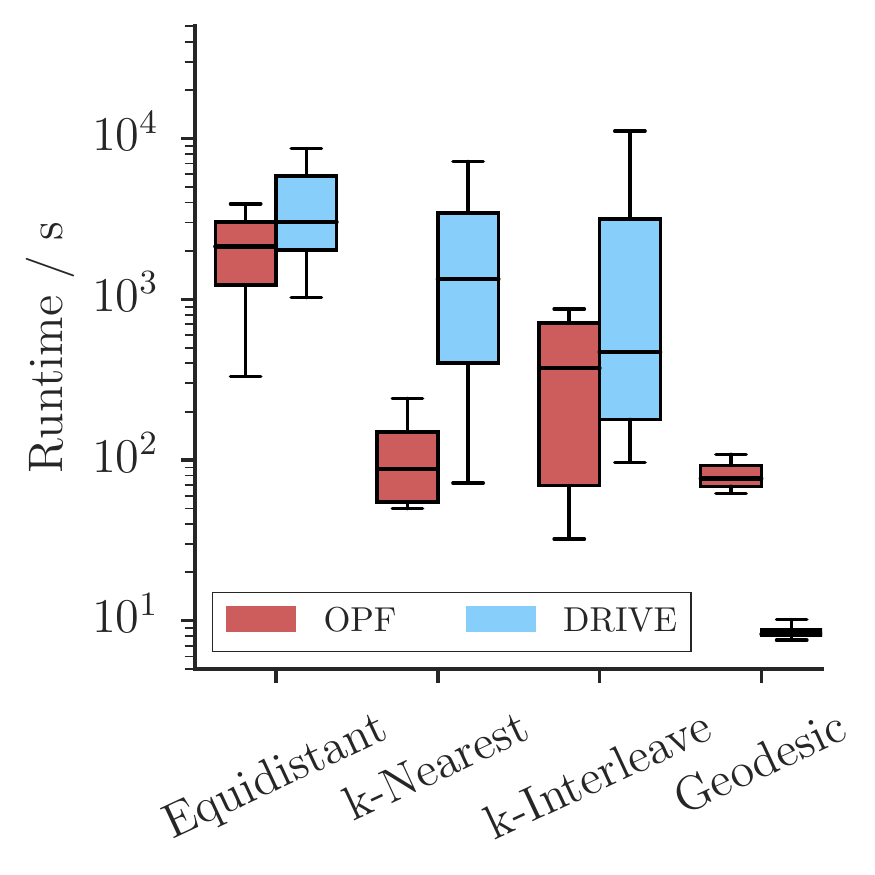}
    }{
      \caption{{\bf Left}: Number of solved instances per
        strategy. The darker bar indicates how often a strategy was
        the \emph{fastest} to solve an instance. {\bf Right}: Runtime
        on solved instances. Strategies with too few solved instances
        are not included. $k$-Nearest and $k$-Interleave are found to
        be the most successful exact strategies.}
      \label{fig:runtime}
    } \ttabbox[0.45\textwidth][]{

    \label{tbl:scores}
    \scriptsize{ 
\begin{tabular}{l|rrrrrr}
\toprule
&\multicolumn{3}{c}{ OPF } & \multicolumn{3}{c}{ DRIVE }\\
 & F1 & (P & R) & F1 & (P & R) \\
\midrule
Maxcomp & 68.5 & (67.7, & 71.9) & 78.7 & (87.2, & 72.1) \\
Geodesic & 76.2 & (69.1, & 85.4) & 80.1 & (86.2, & 75.2) \\
Topocut & - & - & - & 80.1 & (86.4, & 74.9) \\
Exact & 76.2 & (69.1, & 85.4) & 80.1 & (86.2, & 75.2) \\
\bottomrule
\end{tabular}

    }}{\caption{Segmentation scores in terms of \textbf{F1}-score,
      (\textbf{P}recision, \textbf{R}ecall) in \% on the solved
      instances. All approaches outperform the baseline (MaxComp),
      while no significant difference can be found between them.}}
\end{floatrow}
\end{figure}

\begin{figure}[t]
  \floatbox[{\capbeside\thisfloatsetup{capbesideposition={right,top},
capbesidewidth=sidefil}}]{figure}[\FBwidth]
{\caption{Comparison of exact and approximative connectedness: Major
    differences as the one indicated are encountered mainly if
    solutions are competing under the model $\cp{\Xx}{I}$ and thus
    almost equivalent w.r.t. objective value.}
  \label{fig:qual}}
{
 \includegraphics[width=0.68\textwidth]{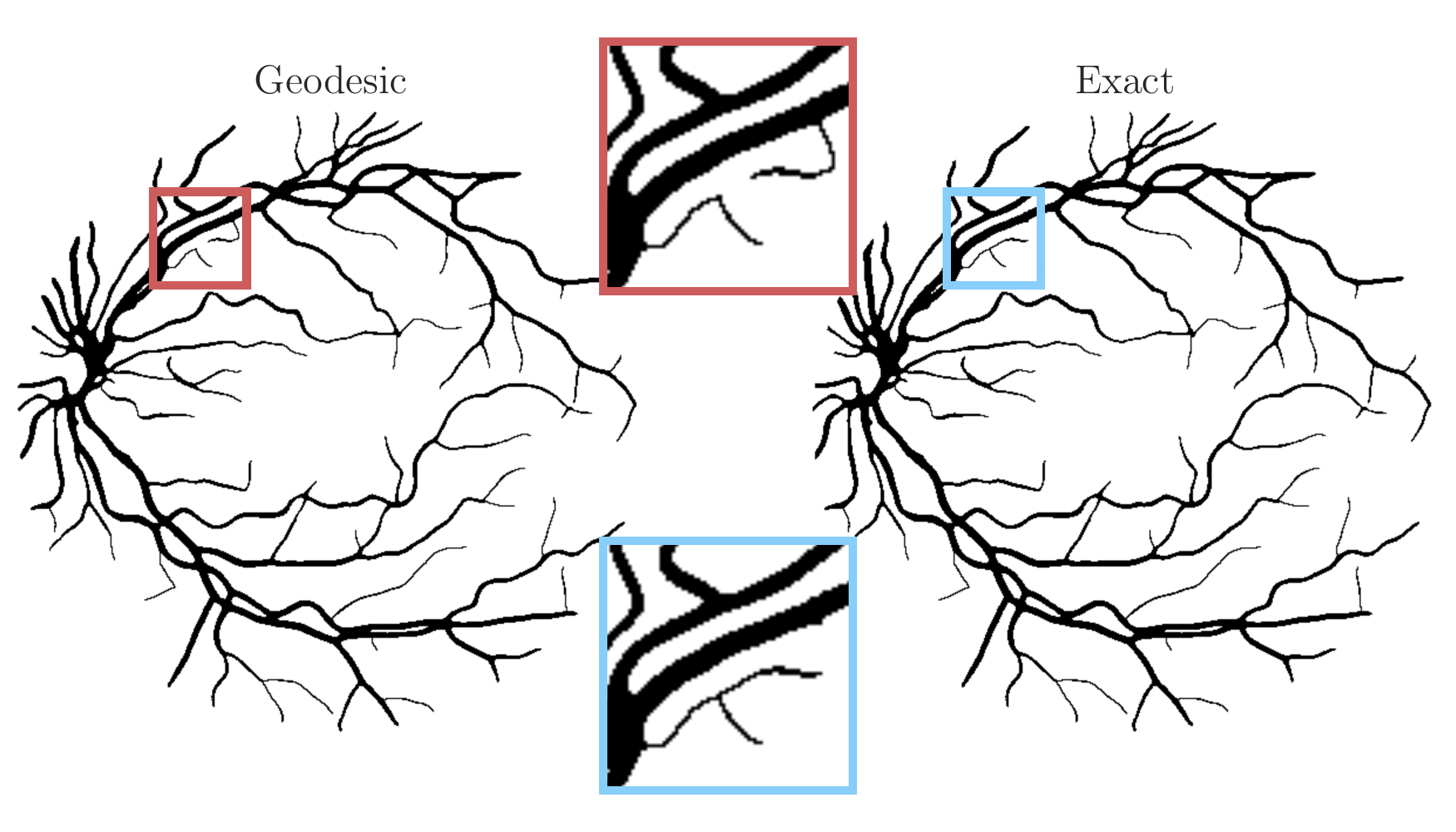}
}
\end{figure}

We compare exact and geodesic tree \ac{mccs} on both datasets. On 2D
images, we additionally compare to the method by~\cite{Chen2011}
called Topocut. As a baseline, we compute the maximum connected
component in the non-constrained solution (Maxcomp). The results are
presented in \figref{fig:runtime} and \tableref{tbl:scores}
(additional information \textit{per instance} is provided in the
supplement). We observe that $6/8$ and $12/20$ instances were solved
to optimality with our propositions, while standard strategies solved
$\leq 1$. $k$-Nearest and $k$-interleave are the two most successful
exact strategies in terms of solved instances and speed. In terms of
segmentation scores, the two heuristics are on par with the exact
algorithm, while all of them outperform the baseline. We find the
geodesic approach to match the exact solution with respect to
objective values in all instances (within a relative difference of
$10^{-4}$), whereas Topocut often obtains slightly lower objective
values than the geodesic approach. A qualitative comparison between an
exact and geodesic solution is presented in \figref{fig:qual}.


\section{Conclusions}

We have shown that exact optimization of the \ac{mccs}, as it is
typical for neural and vascular structure reconstruction tasks,
strongly benefits from the proposed objective-dependent constraints
and the constraint generation strategies. In a first quantitative
comparison between exact and approximative approaches on two datasets,
we found that the geodesic tree formulation is a fast, yet highly
competitive alternative to exact optimization.

While we focussed on large grid-graphs that are most important for
low-level segmentation and reconstruction, we expect that our findings
transfer to \ac{mccs} problems and related \acs{ilp}-based
formulations on sparse graphs, \eg those discussed
in~\cite{Turetken2016,Rempfler2015media,Robben2016,Payer2016}, and
thus consider this a promising direction for future work. Besides, it
will be intersting to investigate the effect of our propositions in the
presence of higher-order terms.

\small{\subsubsection{Acknowledgements.}  With the support of the
  Technische Universit\"at M\"unchen -- Institute for Advanced Study,
  funded by the German Excellence Initiative (and the European Union
  Seventh Framework Programme under grant agreement n 291763).}

\bibliography{library}

\end{document}